%% file: acl_latex.tex
\documentclass[11pt]{article}

\usepackage[final]{acl}

\usepackage{times}
\usepackage{latexsym}
\usepackage{amssymb}
\usepackage{CJKutf8}
\usepackage{textcomp}

\usepackage[T1]{fontenc}

\usepackage[utf8]{inputenc}

\usepackage{microtype}

\usepackage{inconsolata}

\usepackage{graphicx}
\usepackage{amsmath}
\usepackage{tikz}

\usepackage{booktabs}
\usepackage{listings} 
\usepackage{xcolor}  

\usepackage{listings} 

\usepackage{forest}
\useforestlibrary{linguistics}
\forestset{
acl tree/.style={
    for tree={
        parent anchor=south,
        child anchor=north,
        align=center,
        font=\scriptsize,
        l sep=8pt,     
        s sep=3pt,     
        inner sep=2pt,  
        edge path={
            \noexpand\path [draw, \forestoption{edge}]
            (!u.parent anchor) -- ([yshift=-1.5pt].child anchor)\forestoption{edge label};
        },
        before typesetting nodes={
            if content={}{shape=coordinate}{}
        }
    },
    before drawing tree={
        sort by=y,
        for min={tree}{baseline}
    }
}
}

\usepackage{tabularx}   
\usepackage{multirow}   
\usepackage{booktabs}
\usepackage{CJKutf8}
\usepackage{xcolor}
\usepackage{graphicx}

\title{BrailleLLM: Braille Instruction Tuning with Large Language Models for Braille Domain Tasks}

\author{
    Tianyuan Huang$^{1}$\thanks{\,\, Equal contribution.} , 
    Zepeng Zhu$^{1}$\footnotemark[1] , 
    Hangdi Xing$^{2}$\footnotemark[1]\thanks{\,\, Corresponding author.} , 
    Zirui Shao$^{1}$ \\ 
    \textbf{Zhi Yu}$^{1, 4}$\footnotemark[2],
    \textbf{Chaoxiong Yang}$^{1}$, 
    \textbf{Jiaxian He}$^{1}$,
    \textbf{Xiaozhong Liu}$^{3}$,
    \textbf{Jiajun Bu}$^{1}$ \\
    $^{1}$ Zhejiang Key Laboratory of Accessible Perception and Intelligent Systems, \\ \quad Zhejiang University \\ 
    $^{2}$ Alibaba Group \quad
    $^{3}$ Worcester Polytechnic Institute \\
    $^{4}$ Hangzhou High-Tech Zone (Binjiang) Institute of Blockchain and DataSecurity \\
    \texttt{\{huangty, xinghd, yuzhirenzhe\}@zju.edu.cn}
}

\begin{document}
\maketitle
\begin{abstract}

Braille plays a vital role in education and information accessibility for visually impaired individuals. However, Braille information processing faces challenges such as data scarcity and ambiguities in mixed-text contexts. We construct English and Chinese Braille Mixed Datasets (EBMD/CBMD) with mathematical formulas to support diverse Braille domain research, and propose a syntax tree-based augmentation method tailored for Braille data. To address the underperformance of traditional fine-tuning methods in Braille-related tasks, we investigate Braille Knowledge-Based Fine-Tuning (BKFT), which reduces the learning difficulty of Braille contextual features. BrailleLLM employs BKFT via instruction tuning to achieve unified Braille translation, formula-to-Braille conversion, and mixed-text translation. Experiments demonstrate that BKFT achieves significant performance improvements over conventional fine-tuning in Braille translation scenarios. Our open-sourced datasets and methodologies establish a foundation for low-resource multilingual Braille research\footnote{\url{https://github.com/Tianyuan-Huang/BrailleLLM}}.

\end{abstract}

\section{Introduction}
\input{tex_files/1_introduction_v2}

\section{Related Work}
\input{tex_files/2_Related_Work_v2}

\section{Dataset}
\input{tex_files/3_Dataset_v2}

\section{Methodology}
\input{tex_files/4_Methodology_v2}

\section{Experiment}

\input{tex_files/5_Experiment_v2}

\section{Conclusion}
\input{tex_files/6_Conclusion_v2}

\section*{Limitations}
\input{tex_files/7_Limitations_v2}

\section*{Acknowledgements}
\input{tex_files/8_Acknowledgements}

\nocite{Ando2005,andrew2007scalable,rasooli-tetrault-2015}

\bibliography{acl_latex}

\clearpage 
\appendix
\section{Appendix}
\input{tex_files/9_Appendix}

\end{document}

%% file: tex_files/1_introduction_v2.tex
\begin{figure*}[t]
\centering
\includegraphics[width=\textwidth]{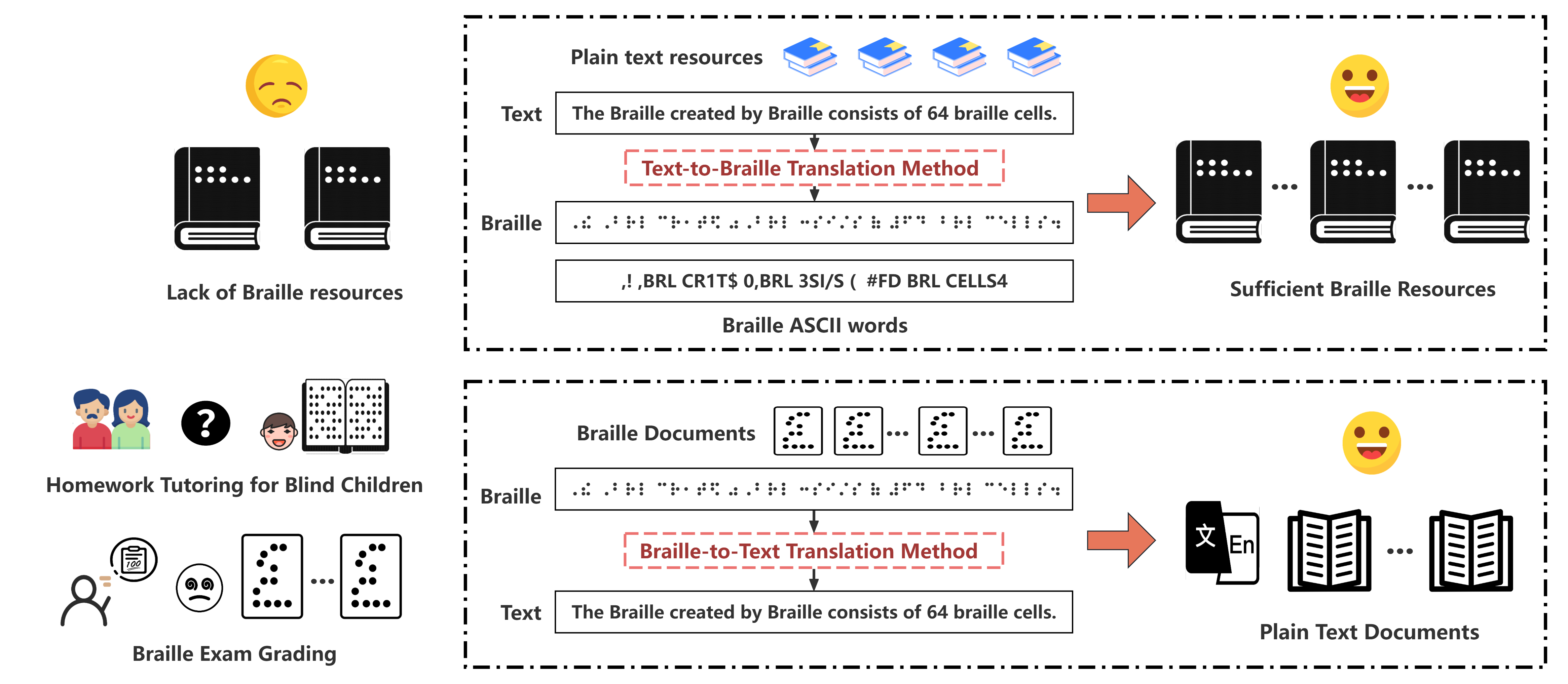}  
\caption{Braille translation methodologies enable the conversion of plain-text resources into Braille or Braille ASCII words (the computational representation of Braille), granting visually impaired populations access to Braille materials. Furthermore, Braille-to-text translation technology enhances educational accessibility by facilitating home tutoring for blind children and empowering Braille school educators to evaluate complex Braille examination papers efficiently.}
\label{fig:fig1}
\end{figure*}

Braille is a tactile writing system essential for the visually impaired, providing a crucial means of accessing textual information. While speech synthesis has improved accessibility, Braille remains vital in areas like education and leisure reading for the blind \cite{johan2022importance,awang2024innovative,kana2024factors}. However, the scarcity of Braille resources and challenges in Braille comprehension have adversely impacted the well-being of the visually impaired community, limiting their access to education. Braille text translation technology offers a promising solution, as illustrated in Fig. \ref{fig:fig1}. Existing Braille research typically focuses on isolated tasks \cite{wang2019cbconv,huang2023translating,yu2023pre}, while visually impaired individuals often need to process documents containing mixed content, including text, formulas, and symbols. Furthermore, the scarcity and limited accessibility of Braille resources further hinder progress in this field.

The primary challenge in mixed-content Braille tasks lies in the inherent ambiguity of Braille-to-text conversion. Braille consists of only 64 distinct Braille cells, typically represented by corresponding Braille ASCII codes in computing systems, which map to multiple plain-text counterparts, including Chinese characters, English words, punctuation, and mathematical symbols. This inherent ambiguity enables a single Braille segment to carry diverse semantic interpretations. For instance, a Chinese Braille segment may correspond to over a dozen Chinese characters. Such a polysemous nature necessitates learning contextual features from extensive training data for robust Braille-Chinese translation. 

Braille translation research has progressed from rule-based and statistical methods to neural machine translation (NMT) \cite{huang2023translating,yu2023pre}, substantially improving the learning capacity for Braille sequence patterns. Nevertheless, studies remain inadequate in addressing the mutual conversion of hybrid-content documents due to Braille translation's inherent complexity and resource limitations. In low-resource scenarios, transferring generic semantic knowledge from large-scale models to domain-specific tasks has demonstrated effectiveness \cite{chowdhery2023palm,touvron2023llama}. 

This paper presents BrailleLLM, a framework that employs instruction-tuned large language models to resolve Braille domain challenges. We construct a comprehensive Braille corpus and design an instruction fine-tuning strategy to enhance model performance. Specifically, we curate task-specific instruction templates and create hybrid Braille-text datasets, including the Chinese Braille Mixed-text Dataset (CBMD) and English Braille Mixed-text Dataset (EBMD). While powerful, LLMs treat Braille as opaque token sequences, forcing them to inefficiently learn its fundamental syntactic rules from scarce data during fine-tuning. We introduce Braille Knowledge-Based Fine-Tuning (BKFT) to overcome this by directly injecting these structural priors. BKFT thus redirects the model's capacity from rediscovering low-level rules to mastering high-level semantic disambiguation and translation.

Our main contributions can be summarized as follows:
\begin{itemize}
    \item We developed the CBMD and EBMD datasets for mixed-text Braille conversion, along with corresponding instruction templates. These resources comprehensively cover various Braille-related tasks.

    \item Based on the unique characteristics of Braille data, we propose a data augmentation method utilizing constituent and dependency syntax trees. This approach effectively increases dataset diversity while maintaining data validity.

    \item We introduce a fine-tuning method based on a Braille prior knowledge base, enabling large general-purpose models to acquire Braille-specific prior knowledge. This method provides a universal foundational approach for encoding Braille data.
    
\end{itemize}

%% file: tex_files/2_Related_Work_v2.tex
\subsection{Braille Information Processing}
Braille information processing research focuses on bidirectional conversion between natural languages and Braille systems. Conventional approaches primarily employ rule-based and statistical models \cite{jiang2002braille,zhang2022design,minghu2000segmentation,wang2017chinese,wang2016chinese,jariwala2017conversion}: pinyin knowledge bases \cite{jiang2002braille} and phonographic mapping tables \cite{wang2017chinese} have driven the development of Chinese-Braille conversion systems, while mathematical formula transformation relies on predefined symbolic rules \cite{jariwala2017conversion,zatserkovnyi2019analysis,egli2009liblouis}. With the evolution of NMT \cite{Cai2019Automatic,Jiang2021End,wang2019cbconv,huang2023translating,yu2023pre}, researchers have developed end-to-end Braille conversion models. Although NMT demonstrates adaptability in multi-lingual scenarios \cite{vaswani2017attention,zhou2022confidence,hu2021deep,lu2022learning}, its dependence on large-scale annotated data inherently conflicts with the scarcity of Braille corpora. Current studies predominantly concentrate on single-content conversion, yet fail to address the critical challenge of mixed-content processing faced by visually impaired communities, resulting in limited applicability of existing methods in real-world scenarios.

\subsection{Low resource translation}
Research on low-resource translation primarily addresses data scarcity and encoding complexity through transfer learning, meta-learning, and data augmentation. Transfer learning enhances model adaptability via pre-training and fine-tuning paradigms \cite{di2017monolingual,qi2018and,dong2023transfer,gao2024novel}. Meta-learning approaches optimize parameter initialization and generalization through cross-domain knowledge sharing \cite{li2020metamt,park2020unsupervised}. Data augmentation techniques synthesize parallel corpora via back-translation \cite{sennrich2015improving,bawden2019university,sanchez2020english}, syntax-rule generation \cite{lucas2024grammar}, and bilingual templates \cite{li2024bilingual}, with syntax-tree strategies offering novel paradigms for morphologically complex languages \cite{lucas2024grammar}. However, existing Braille translation studies have not effectively leveraged knowledge transfer methodologies, and current data augmentation approaches lack tailored solutions for Braille-specific data characteristics.

\subsection{Large Language Models}
LLMs have revolutionized natural language processing through the self-supervised pretraining paradigm based on Transformer architectures. Pioneering works like BERT \cite{devlin2018bert} and GPT series \cite{radford2018improving,radford2019language,brown2020language} models demonstrated the efficacy of the pretrain-finetune paradigm in cross-task transfer. Continued model scaling and architectural innovations have enhanced LLMs' generalization capabilities \cite{chowdhery2023palm,touvron2023llama}. Despite contemporary LLMs demonstrate general task-solving capabilities \cite{liu2024deepseek,achiam2023gpt}, their applications in specialized domains remain constrained by domain-specific knowledge gaps. Existing domain-adapted solutions (e.g., Med-PaLM \cite{singhal2023large} and LawLLM \cite{shu2024lawllm}) enhance task performance through instruction tuning and knowledge integration. Nevertheless, current general-purpose LLMs exhibit inadequate Braille processing capabilities and lack Braille-specific adaptations.

%% file: tex_files/3_Dataset_v2.tex
We present the first comprehensive multi-lingual Braille dataset suite for Chinese and English Braille processing tasks, addressing the critical data scarcity challenge in Braille NLP. Our dataset construction integrates three key components: (1) a Chinese Braille Dataset (CBD), (2) a Chinese Braille Mixed-text Dataset (CBMD), and (3) an English Braille Mixed-text Dataset (EBMD). To enable instruction tuning for large language models (LLMs), we generate task-specific datasets through diverse instruction templates. Given the high annotation cost of Chinese Braille, we propose a novel syntax-aware data augmentation method tailored to its linguistic properties.

\subsection{Data Collection}
The collection of Braille datasets has faced significant challenges, especially in acquiring parallel datasets between Plain text and Braille texts. Blind schools and Braille libraries often use Braille, but structured parallel data is difficult to obtain and verify. As a result, we first collected plain texts, then used conversion tools to generate an initial Braille version, followed by manual correction and annotation.
For the CBD, we selected three months of standardized Chinese news articles from the People’s Daily. The CBMD construction focuses on STEM education needs, utilizing junior and senior high school mathematics documents containing Chinese text and mathematical formulas. For EBMD, we adapt the open-source NuminaMath-CoT \footnote{https://huggingface.co/datasets/AI-MO/NuminaMath-CoT} dataset, comprising English mathematical problems from online exams and forums.
Raw texts undergo rigorous preprocessing, including deduplication, format normalization using LaTeX standardization for mathematical formulas, sanitization to filter non-standard characters, and error correction through rule-based verification. Dataset statistics and sample structures are detailed in Table \ref{tab:table1}. 

\begin{table*}[t]  
\centering  
\begin{tabular}{lcc}  
\toprule  
\textbf{Data type} & \textbf{Data volume} & \textbf{Data Example} \\  
\midrule  
Pure Chinese & 11000 & \begin{CJK}{UTF8}{gbsn} 该增长水平使手机市场成为当今最活跃的商品市场之一。 \end{CJK} \\ 
Mixed Chinese & 11000 &\begin{CJK}{UTF8}{gbsn} 答：图1中阴影部分面积为：\$\textbackslash frac\{3\}\{m+1\}\$ 平方米 \end{CJK}  \\  
Mixed English & 11000 &  Suppose that \$g(x)=5x-3\$. What is \$g\textasciicircum\{-1\}(g\textasciicircum\{-1\}(14))\$?  \\  
\bottomrule  
\end{tabular}

\caption{Data sample and statistics.}  
\label{tab:table1}
\end{table*}

\subsection{Data Annotation}

The acquisition of initial Braille was obtained by utilizing text-to-Braille and formula-to-Braille conversion libraries through concatenation to generate hybrid Braille. For Chinese mixed text-to-Braille conversion, dedicated APIs corresponding to Chinese and mathematical formulas from the China Braille Digital Platform \footnote{http://www.braille.org.cn/} were employed. English hybrid Braille generation utilized the English and mathematical formula conversion interfaces in the liblouis library \footnote{https://github.com/liblouis/liblouis}.

Despite the automatic conversion, errors and inconsistencies remained, including incorrect word segmentation, tonal marking errors, and punctuation issues in Chinese Braille, as well as translation errors in both English words and formulas. We identified various errors and compiled detailed annotation guidelines for both Chinese and English Braille. A team of 10 Braille domain experts and AI specialists was tasked with the annotation and quality control process. Multiple rounds of validation were performed to ensure the accuracy of the final datasets. The resulting datasets, CBD, CBMD, and EBMD, provide word-level alignment between plain text and Braille text.

\begin{figure*}[t]  
\centering
\includegraphics[width=\linewidth]{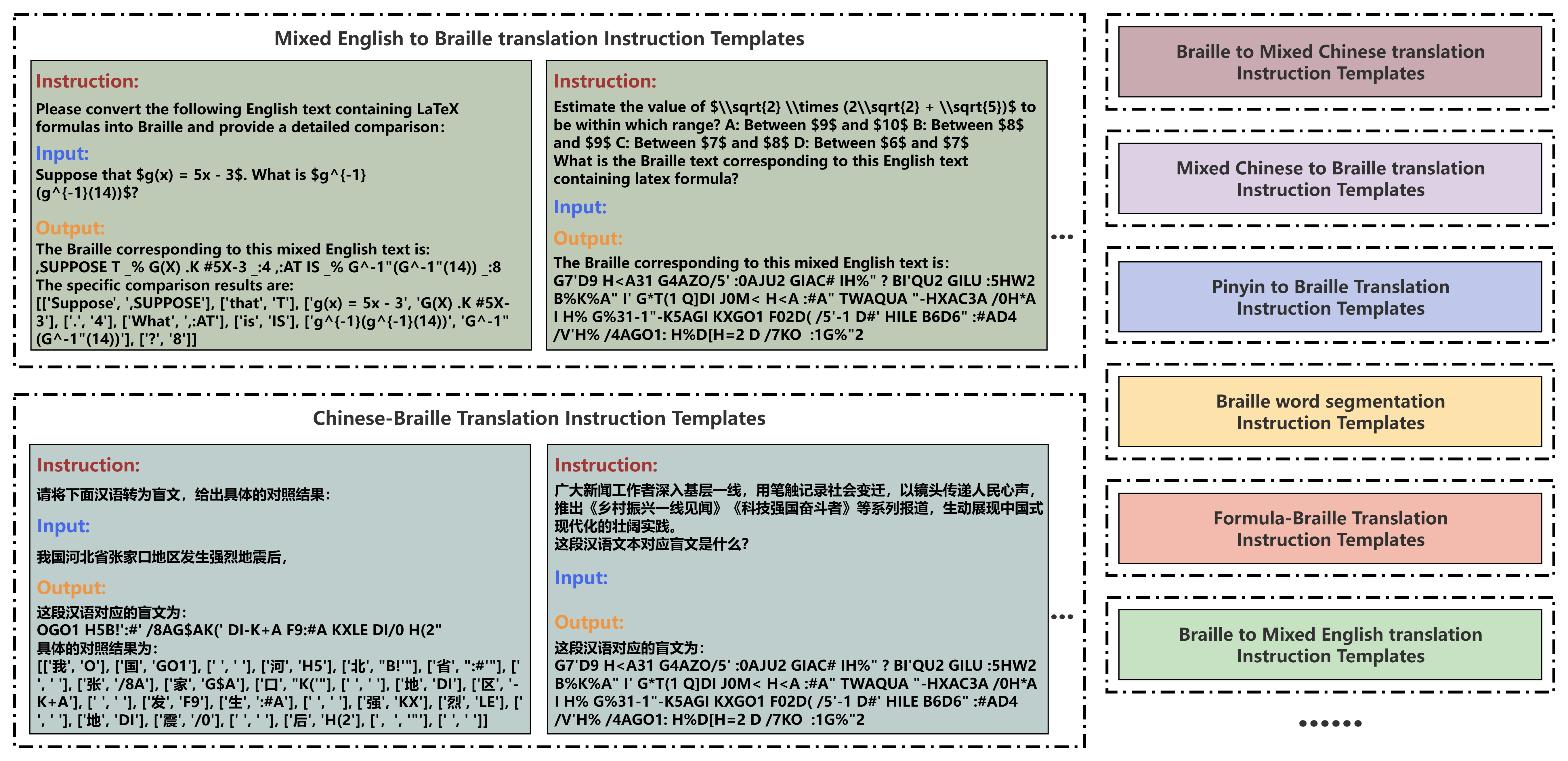}  
\caption{Instruction template samples.}
\label{fig:fig2}
\end{figure*}

To support additional Braille domain tasks, such as Braille word segmentation and conversion between Braille and Pinyin, we expanded the datasets to include specific task data, as illustrated in Table \ref{tab:table2}. Furthermore, we developed dozens of instruction templates and corresponding instruction datasets, as shown in Figure \ref{fig:fig2}.

\begin{table*}[t]  
\centering  
\setlength\tabcolsep{2.7pt}
\begin{tabular}{lcc}  
\toprule  
\textbf{Braille task type} & \textbf{Source sequence} & \textbf{Target sequence} \\  
\midrule  
Chinese-to-Braille & \begin{CJK}{UTF8}{gbsn} 经济的快速发展\end{CJK} &  G*AGI D KYSU F9/V' \\ 
Pinyin-to-Braille & Jing1ji4defa1zhan3 & G*AGI D KYSU F9/V'\\  
Formula-to-Braille & \$frac{1}{4}x=15\$ & \#A4;X 7\#AE  \\  
Braille word segmentation & G*AGIDKYSU F9/V' & G*AGI D KYSU F9/V'   \\ 
Mixed Chinese to Braille & \begin{CJK}{UTF8}{gbsn} 故答案为：\$y\$ \end{CJK} & GU D91V W- \#Y \\  
Mixed English to Braille  & Suppose that \$g(x) = 5x - 3\$. &  ,SUPPOSE T \_ \% G(X) .K \#5X-3  \\ 
\bottomrule  
\end{tabular} 
\caption{Examples of some Braille task data.}  
\label{tab:table2}  
\end{table*}

\subsection{Braille Data Augmentation}

The creation of Chinese Braille datasets is notably more complex and incurs significantly higher verification costs compared to English Braille, due to the highly ambiguous character representations in Chinese Braille, whereas English words and Braille exhibit clear correspondences. To address this challenge, we investigate data augmentation methods tailored to the semantic composition characteristics of Chinese Braille texts.

\begin{figure}
\centering
\includegraphics[width=\linewidth]{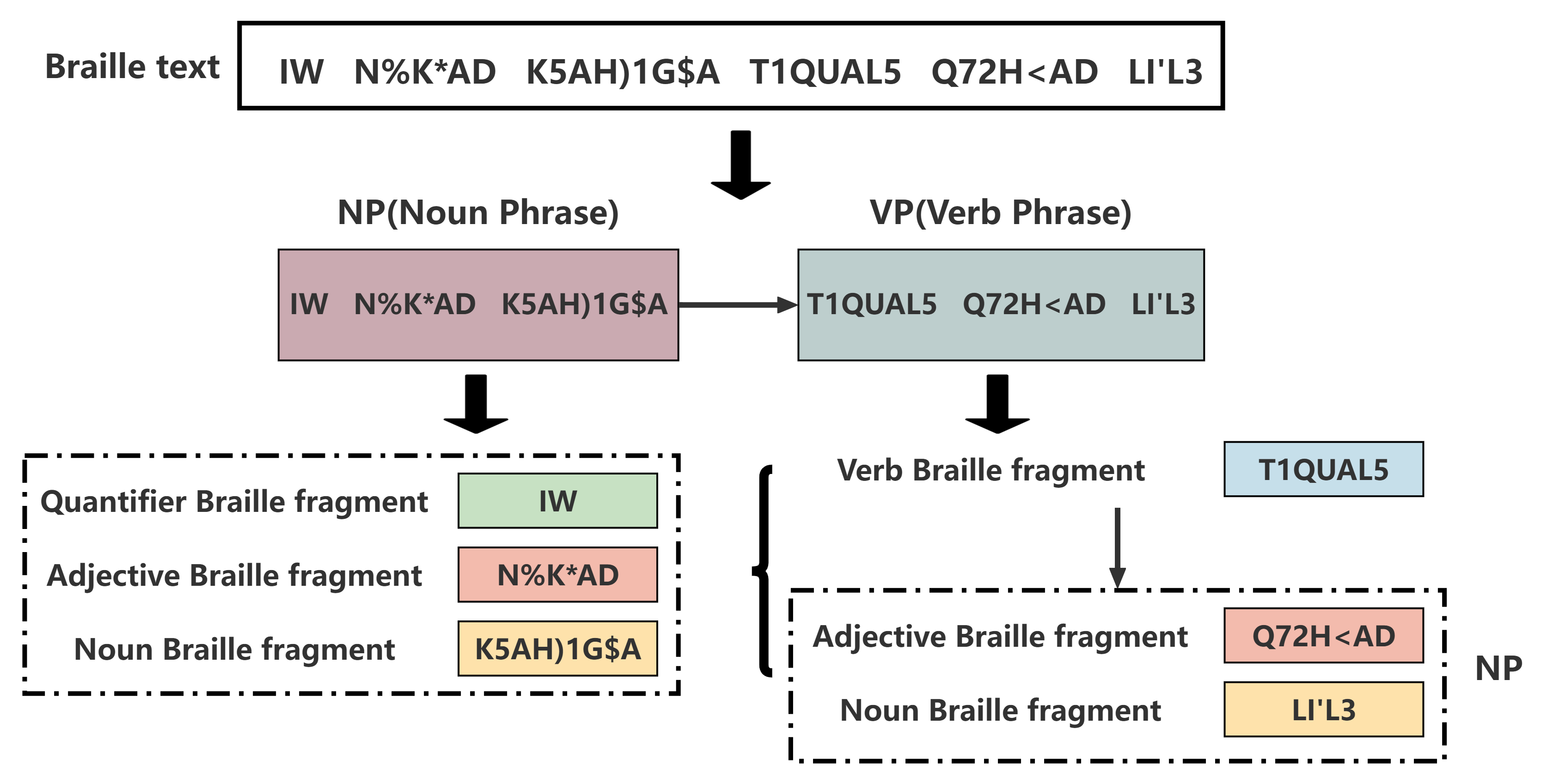}
\caption{Example of component analysis of Braille text.}
\label{fig:fig3}
\end{figure}

By performing constituency syntax tree parsing on Braille texts, we categorize Braille segments into dozens of attribute types, including quantifier attributes, pronoun attributes, and location attributes. For example, the Braille sequence "IW N\%K*AD K5AH)1G\$A T1QUAL5 Q72H<AD LI'L3" corresponds to the constituency syntax tree shown in Figure \ref{fig:fig3}. The NP (noun phrase) component comprises a quantifier Braille segment "IW", an adjective Braille segment "N\%K*AD", and a noun Braille segment "K5AH)1G\$A".  The interdependencies among Braille segments of different attributes within the syntax tree ensure the linguistic validity of Braille texts.  

We construct a knowledge base encompassing various Braille attribute segments from existing Braille data, partially exemplified in Table \ref{tab:table3_attri}. During the data augmentation phase, new Braille sequences are synthesized by leveraging the dependency syntax trees of original Braille sequences and replacing constituent segments with semantically compatible counterparts from the knowledge base. Segment similarity matching algorithm is employed to ensure the syntactic coherence of the reconstructed sequences.

\begin{table*}[t]  
\centering  
\setlength\tabcolsep{4pt}
\begin{tabular}{lcc}  
\toprule  
\textbf{Attributes of the Braille fragments} & \textbf{Number of fragments} & \textbf{Fragment examples} \\  
\midrule  
Verb Braille fragment & 11048 & \%'-QUA | T(1SU \\ 
Personal name Braille fragment & 6717 & K4'K*2H5 | :AD51-:A \\  
Quantifier Braille fragment & 3658 & B9A:1SVASW | SVAF0/AR2\\  
Place name Braille fragment & 3482 & H<AHIALV1 | TU'KUMV2SATV' \\  
\bottomrule  
\end{tabular} 
\caption{Example of partial Braille fragments attribute knowledge base.}  
\label{tab:table3_attri} 
\end{table*}

%% file: tex_files/4_Methodology_v2.tex
\begin{figure*}[t]  
\centering
\includegraphics[width=\textwidth]{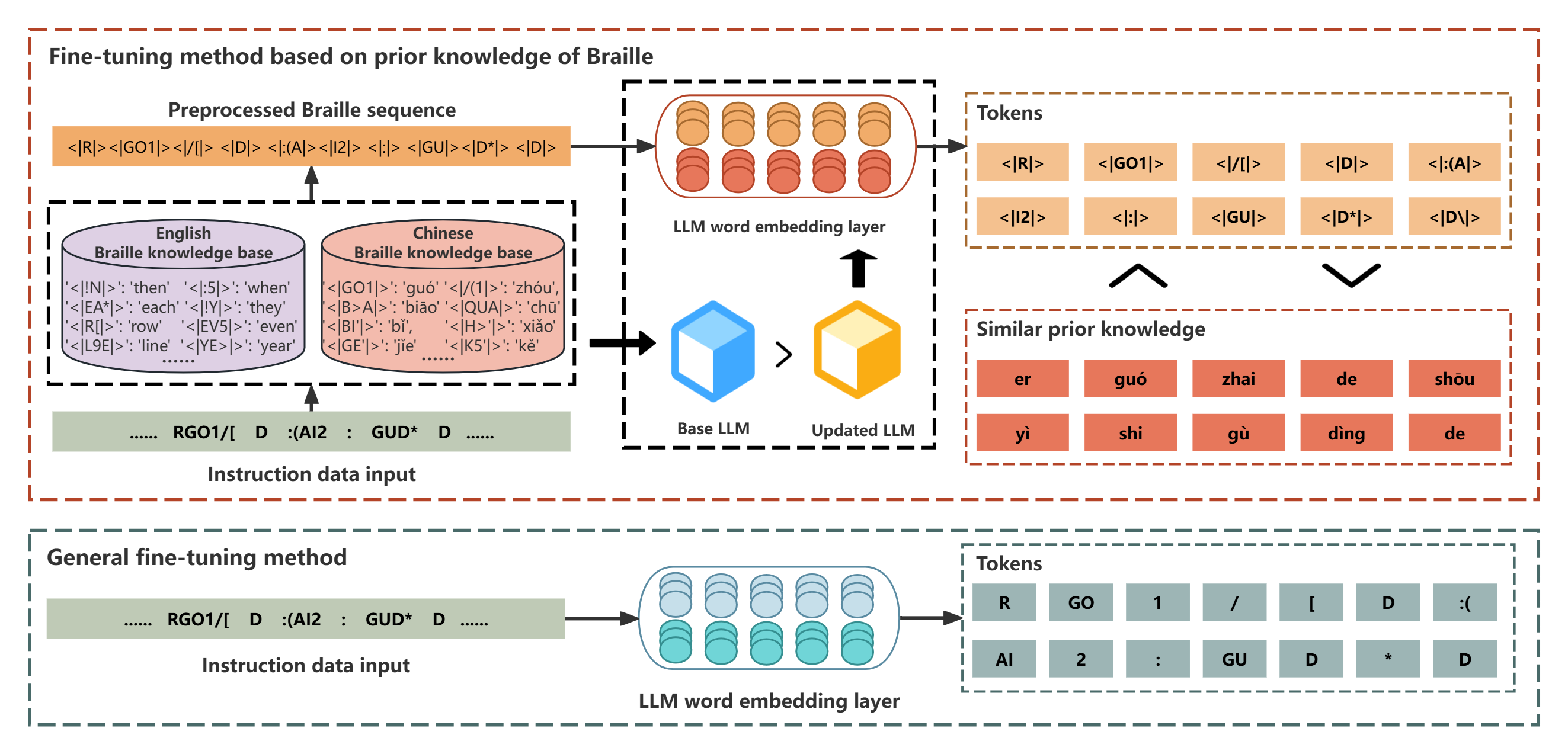}  
\caption{Comparison of fine-tuning based on Braille prior knowledge with conventional approaches. This study constructs a mapping knowledge base between Chinese/English Braille segments and their semantically equivalent words, enabling more rational segmentation of Braille sequences. Simultaneously, Braille segment embeddings are initialized with high-frequency equivalent word embeddings, which alleviates the model's comprehension difficulty of Braille patterns during fine-tuning.}
\label{fig:fig4}
\end{figure*}

This study presents a fine-tuning framework for the Braille domain based on large-scale language models (LLMs), which integrates Braille-specific prior knowledge into the fine-tuning process to address the limitations of existing LLMs in handling Braille tasks, as illustrated in Fig. \ref{fig:fig4}. Our approach introduces two distinct methods, CBKFT and EBKFT, tailored for Chinese and English Braille respectively, leveraging linguistic characteristics of each language. The unified fine-tuning procedure combines these methods to enhance the Braille comprehension and generation capabilities of the model.

\subsection{CBKFT: Fine-Tuning Based on Chinese Braille Prior Knowledge}
Chinese Braille is a syllable-based system, where each Braille fragment may correspond to multiple Chinese characters, but typically represents a single pinyin syllable (e.g., \texttt{HV2} $\rightarrow$ \texttt{hàn}). Since LLMs lack inherent understanding of Braille symbols, CBKFT bridges this gap by mapping Braille fragments to their phonetic syllables. We construct a \textit{Chinese Braille Prior Knowledge Base} $\mathcal{K}_C = \{ (b_i, p_j) \}$, where $b_i \in \mathcal{B}_C$ denotes a Braille fragment and $p_j \in \mathcal{P}$ represents its corresponding Pinyin syllable. Here, $\mathcal{B}_C$ is the set of 1,152 Chinese Braille fragments, and $\mathcal{P}$ is the Pinyin syllable inventory.

Given an input Braille sequence $S_C = \{b_1, b_2, \ldots, b_n\}$, we tokenize $S_C$ into Braille-specific tokens (e.g., \texttt{<|G*A|>}, \texttt{<|GI|>}, \texttt{<|F9|>}, \texttt{<|/V'|>}) and map each token to its syllable $p_j$ via $\mathcal{K}_C$. Let $\mathcal{T}_p \subseteq \mathcal{V}$ denote the subset of token IDs in the LLM's vocabulary that correspond to syllable $p_j$, where $\mathcal{V}$ is the full vocabulary. The embedding of Braille token $b_i$, denoted $\mathbf{e}_{b_i}$, is initialized as the mean of embeddings for all tokens in $\mathcal{T}_p$:

\begin{equation}
\mathbf{e}_{b_i} = \frac{1}{|\mathcal{T}_p|} \sum_{t \in \mathcal{T}_p} \mathbf{E}_t,
\end{equation}

where $\mathbf{E} \in \mathbb{R}^{|\mathcal{V}| \times d}$ is the LLM's embedding matrix, and $d$ is the embedding dimension. This syllable-aware initialization injects phonetic priors into Braille token embeddings, enabling the model to leverage shared phonological features across homophonic characters during fine-tuning. Formally: $\mathbf{E}[b_i] \leftarrow \mathbf{e}_{b_i}.$ 


\subsection{EBKFT: Fine-Tuning Based on English Braille Prior Knowledge}
English Braille fragments map directly to whole words rather than syllables (e.g., \texttt{L1/} $\rightarrow$ \texttt{least}). EBKFT utilizes an \textit{English Braille Prior Knowledge Base} $\mathcal{K}_E = \{ (b'_i, w_j) \}$, where $b'_i \in \mathcal{B}_E$ is an English Braille fragment and $w_j \in \mathcal{W}$ is its corresponding word, with $\mathcal{B}_E$ containing 6,000 common English Braille fragments. For an input Braille sequence $S_E = \{b'_1, b'_2, \ldots, b'_m\}$, each token (e.g., \texttt{<|L1/|>}) is mapped to its word $w_j$ via $\mathcal{K}_E$. The embedding of $b'_i$, $\mathbf{e}_{b'_i}$, is initialized by cloning the embedding of $w_j$: $\mathbf{E}[b'_i] \leftarrow \mathbf{E}[w_j],$ where $\mathbf{E}[w_j]$ is the pretrained embedding of word $w_j$. This direct lexical transfer preserves semantic and syntactic knowledge from the LLM's original vocabulary.


\subsection{Model Fine-Tuning Procedure}
The fine-tuning procedure unifies the Braille-specific initialization strategies of CBKFT and EBKFT with end-to-end training to adapt the LLM for Braille understanding and generation. Given a parallel dataset $\mathcal{D} = \mathcal{D}_C \cup \mathcal{D}_E$ containing Braille sequences paired with their textual counterparts, the training objective minimizes the negative log-likelihood of generating target text $Y = \{y_1, \dots, y_T\}$ conditioned on Braille input $S$. The procedure involves three key phases: Braille tokenization and embedding initialization, contextual adaptation, and cross-lingual joint optimization.

For each Braille sequence $S$, we first apply language-specific tokenization rules to segment it into Braille fragments (e.g., \texttt{<|G*A|>} for Chinese or \texttt{<|L1/|>} for English). Each fragment $b_i$ is mapped to its linguistic counterpart (Pinyin syllable $p_j$ or word $w_j$) via the prior knowledge base $\mathcal{K}_C$ or $\mathcal{K}_E$. The token embedding $\mathbf{e}_{b_i}$ is initialized using the procedure described for CBKFT or EBKFT, ensuring $\mathbf{E}[b_i]$ encodes syllable-level or word-level semantics before training.

During training, the model processes the initialized Braille embeddings alongside standard text tokens, enabling it to learn contextual relationships between Braille fragments and their textual meanings. The embeddings of Braille tokens are updated dynamically through backpropagation, guided by the loss $\mathcal{L}(\theta)$. To enhance cross-lingual generalization, we interleave batches from $\mathcal{D}_C$ and $\mathcal{D}_E$, forcing the model to distinguish between Chinese and English Braille patterns while sharing underlying linguistic knowledge. This phased approach ensures that the model first grounds Braille tokens in linguistic priors, then refines their representations through task-specific contextual learning, ultimately achieving robust performance on low-resource Braille tasks.

%% file: tex_files/5_Experiment_v2.tex
\subsection{Implementation Details}
In this study, we used the pre-trained large language model Qwen2.5 7B as the base model, and proposed CBKFT and EBKFT fine-tuning methods based on Braille prior knowledge for Chinese and English Braille tasks, respectively. Comparable baseline models including LLama3 8B, GLM4 9B, and Qwen2.5 7B of similar scales were fine-tuned for performance comparison. All models were fine-tuned using LoRA adapters for efficient parameter updating, with a rank of 8, and the AdamW optimizer was used for training. The initial learning rate was set to 1e-4, and the maximum sequence length was fixed at 1024. To evaluate model performance on Braille domain tasks, we employed multiple metrics: sacreBLEU \footnote{https://github.com/mjpost/sacreBLEU} and chrF++ for translation quality assessment, CER (Character Error Rate) and TER (Translation Error Rate) for error rate analysis, along with BERTScore \cite{zhang2019bertscore} for semantic similarity evaluation between  the model's generated answers and the reference answers. All experiments were conducted on a computing platform equipped with two NVIDIA A40 GPUs.

\subsection{Performance of BKFT}
The fine-tuning method based on Braille prior knowledge enables large language models to comprehend Braille sequences better, thereby enhancing Braille-to-text translation performance. Tables \ref{tab:table4} present the performance of various models under different fine-tuning data sizes for Braille-to-Chinese translation tasks. BrailleLLM achieves the best performance across all data scales, attaining a CER of only 0.0542 with 10,000 fine-tuning samples, demonstrating low character error rates in the generated Chinese text. The results indicate that the performance improvement is more pronounced with smaller fine-tuning datasets. With only 3,000 fine-tuning samples, our method achieves an 18.96 BLEU score improvement compared to approaches without BKFT. This suggests that the BKFT method can effectively learn Braille sequence features even in low-resource scenarios. The comparison against the Qwen2.5-7B baseline, which undergoes only conventional fine-tuning, effectively ablates our BKFT method and validates its significant contribution to the performance improvement.

\begin{table*} 
\centering  
 
\setlength\tabcolsep{7pt}
\begin{tabular}{lccccccccc}  
\toprule  
 & \multicolumn{3}{c}{sacreBLEU $\uparrow$} & \multicolumn{3}{c}{chrF++ $\uparrow$} & \multicolumn{3}{c}{CER $\downarrow$}\\  
 \textbf{Model} & 3K & 6k & 10K &  3K & 6k & 10K &  3K & 6k & 10K \\
\midrule  
GLM4 9B & 51.04 & 78.78 & 84.89 & 49.16 & 66.15 & 73.33 & 0.5670 & 0.1422 & 0.0932  \\ 
Llama3 8B & 52.77 & 70.29 & 77.48 & 42.21 & 58.50 & 68.07 & 0.3372 & 0.2250 & 0.0889  \\ 
Qwen2.5-7B & 63.19 & 77.95 & 85.52 & 50.39 & 65.63 & 74.60 & 0.2660 & 0.1509 & 0.0870  \\ 
BrailleLLM & \textbf{82.15} & \textbf{88.37} & \textbf{91.15} & \textbf{69.69} & \textbf{77.82} & \textbf{82.18} & \textbf{0.1180} & \textbf{0.0739} & \textbf{0.0542}  \\ 
  
\bottomrule  
\end{tabular}

\caption{Performance of various models under different fine-tuning data quantities for Braille-to-Chinese translation tasks. Note: BrailleLLM (Qwen2.5 7B + BKFT).}  
\label{tab:table4}  
\end{table*}

\begin{table*} 
\centering  
 
\setlength\tabcolsep{7pt}
\begin{tabular}{lccccccccc}  
\toprule  
 & \multicolumn{3}{c}{sacreBLEU $\uparrow$} & \multicolumn{3}{c}{chrF++ $\uparrow$} & \multicolumn{3}{c}{TER $\downarrow$}\\  
 \textbf{Model} & 3K & 6k & 10K &  3K & 6k & 10K &  3K & 6k & 10K \\
\midrule  
GLM4 9B & 77.74 & 86.00 & 87.03 & 76.39 & 82.91 & 86.90 & 20.04 & 11.60 & 12.77  \\ 
Llama3 8B & 78.80 & 84.90 & 88.21 & 76.51 & 81.62 & 85.85 & 18.51 & 14.72 & 10.63  \\ 
Qwen2.5-7B & 76.92 & 85.85 & 88.57 & 73.50 & 82.85 & 86.27 & 24.78 & 12.18 & 11.09   \\ 
BrailleLLM & \textbf{81.85} & \textbf{87.52} & \textbf{89.64} & \textbf{79.19} & \textbf{84.60} & \textbf{87.39} & \textbf{16.87} & \textbf{10.31} & \textbf{9.95}  \\ 
  
\bottomrule  
\end{tabular}
\caption{Performance of different models in the applied scenario of mixed English Braille-to-plaintext translation.}  
\label{tab:table5} 
\end{table*}

\subsection{Braille Translation in Application Scenarios}
Due to spatial perception challenges, visually impaired individuals often make Braille writing errors such as missing dots, extra dots, and other inaccuracies when embossing six-dot Braille characters. Consequently, Braille in real-world applications typically contains such errors and consists of mixed elements, including numbers, punctuation, and mathematical symbols. To further evaluate the effectiveness of the BKFT method, we integrated common writing errors into the fine-tuning data. As shown in Table \ref{tab:table5}, our approach consistently outperforms baseline models on the mixed English Braille-to-text conversion task with erroneous inputs. This demonstrates that BrailleLLM acquires a more robust understanding of imperfect Braille during fine-tuning, highlighting its practical applicability in real-world scenarios.

\subsection{Auxiliary Performance of Braille Data Augmentation}
To evaluate the effectiveness of Braille data augmentation method, we compared it with two commonly used data augmentation strategies: noise injection-based adversarial generation and segment replacement-based adversarial generation. The noise injection method randomly deleted or inserted 15\% of corresponding Braille-text pairs, while the fragment-replacement method replaced fragments within Braille-text pairs. After augmenting 3,000 Braille-Chinese parallel fine-tuning samples, the experimental results are shown in Table \ref{tab:table10}. Our proposed syntax tree-based Braille augmentation method achieved the best auxiliary performance, demonstrating that our approach can effectively enhance the diversity of Braille data and provide more Braille-specific features for the fine-tuning process.

\begin{table} 
\centering  
\setlength\tabcolsep{6pt}
\begin{tabular}{lccccc}  
\toprule  
\textbf{Method} & Chinese-to-Braille  \\  
\midrule  
None & 79.49  \\ 
Noise Injection & 85.71  \\  
Fragment Replacement & 83.19   \\ 
Ours &  \textbf{87.32} \\   
\bottomrule  
\end{tabular}
\caption{Auxiliary performance of different data augmentation methods.}  
\label{tab:table10}
\end{table}

\subsection{Evaluation of BrailleLLM's Question-Answering Capability}

For Braille-related tasks, user input formats are often inconsistent, requiring the model to comprehend the question before generating a response. To evaluate BrailleLLM's question-answering capability, we constructed 1,000 question-answer pairs covering various Braille tasks and assessed the semantic similarity between the model's responses and reference answers using BERTScore, as shown in Table \ref{tab:table11}. The results demonstrate that the model's answers exhibit high semantic alignment with the expected outcomes, proving its effectiveness in response generation and task processing.

\begin{table}  
\centering  
\setlength\tabcolsep{6pt}
\begin{tabular}{lccccc}  
\toprule  
\textbf{Method} & Text-to-Braille & Braille-to-Text \\  
\midrule  
English QA & 0.9605 & 0.9543 \\ 
Chinese QA & 0.9487 & 0.9435 \\  
\bottomrule  
\end{tabular}
\caption{Performance of BrailleLLM on question answering tasks.} 
\label{tab:table11}
\end{table}

Furthermore, we compared BrailleLLM's performance on Braille tasks with general-purpose conversational LLMs, including GPT-4 and Claude 3.5. Current general models exhibit limited capability in processing Chinese Braille. Table \ref{tab:table12} displays model performance on mixed English Braille-to-plaintext translation tasks. BrailleLLM demonstrates a significant performance advantage over existing conversational models in Braille-related tasks.

\begin{table}  
\centering  
\setlength\tabcolsep{6pt}
\begin{tabular}{lccccc}  
\toprule  
\textbf{Model} & sacreBLEU & chrF++ \\  
\midrule  
GPT-4 & 48.14 & 38.69 \\ 
Claude3.5 & 22.76 & 18.50 \\  
BrailleLLM & \textbf{89.94} & \textbf{87.39} \\
\bottomrule  
\end{tabular}

\caption{Comparative performance of BrailleLLM versus general-purpose LLMs on Braille-specific tasks.} 
\label{tab:table12}
\end{table}

%% file: tex_files/6_Conclusion_v2.tex
The paper proposes BrailleLLM, the first multi-task large language model for Braille processing, which enables bidirectional conversion between Braille and mixed-content texts containing formulas and regular text. We construct and open-source a large-scale annotated dataset covering Chinese and English Braille, mitigating the long-standing data scarcity issue in Braille research. An efficient fine-tuning approach tailored for Braille data is proposed to perform instruction tuning with LLMs across various Braille tasks. The fine-tuned model demonstrates superior Braille translation performance and exhibits Braille processing capabilities absent in existing general-purpose conversational LLMs. This work not only enhances understanding of general-purpose models in the Braille domain but also lays the groundwork for future exploration.

%% file: tex_files/7_Limitations_v2.tex
Although the proposed Braille dataset and fine-tuning method show significant effectiveness for Braille tasks, we acknowledge that Braille is often represented in image form, which limits the method's applicability. Future work could expand the dataset to include more forms of Braille input, ensuring better Braille information processing services.

%% file: tex_files/8_Acknowledgements.tex
This work is supported by the National Natural Science, Foundation of China (Grant No. 62372408).

%% file: tex_files/9_Appendix.tex
\subsection{Data Validation Pipeline}

To ensure the fidelity of our bilingual Braille dataset, we designed a rigorous validation pipeline executed by Braille experts. This process systematically verifies correctness across five stages, from low-level format integrity to high-level semantic coherence. Table~\ref{tab:validation_pipeline} outlines the key aspects of each stage.

\begin{table}[h!] 
\centering
\begin{tabularx}{\columnwidth}{l X}
\toprule
Stage & Key Aspects \\
\midrule
Automated Check 
& Checks for invalid Braille ASCII codes and verifies standard LaTeX formatting. \\
\addlinespace 
Unit Accuracy 
& Verifies the correct mapping of individual plaintext elements (e.g., Chinese characters, words, formulas) to Braille, including spelling and tone mark accuracy. \\
\addlinespace
Rule Validation
& Ensures adherence to Braille language rules, such as correct Braille word segmentation, abbreviated tone marks, and contractions. \\
\addlinespace
Cross-Review
& Involves independent review by a different expert on a 20\% data sample to ensure high Inter-Annotator Agreement. \\
\addlinespace
Fluency Review
& Involves a tactile read-through of the Braille to check for natural transitions between text and formulas, and to ensure semantic coherence and absence of contextual ambiguity. \\
\bottomrule
\end{tabularx}
\caption{The Five-Stage Data Validation Pipeline.}
\label{tab:validation_pipeline}
\end{table}

\subsection{Qualitative Analysis of Error Patterns}

To gain deeper insights into the superior performance of BrailleLLM on complex Braille translation tasks, we conduct a qualitative analysis of its error patterns. The core challenge in Braille translation stems from its inherent ambiguity: a single Braille cell or sequence can map to diverse elements, such as characters from different languages, digits, punctuation, or mathematical symbols. This polysemy necessitates a strong contextual understanding of Braille. As illustrated in Table \ref{tab:qualitative_analysis}, the baseline model—a traditionally fine-tuned Qwen2.5-7B—exhibits typical error patterns arising from this ambiguity due to its lack of sufficient Braille-specific prior knowledge. For instance, in Case 1, the baseline model incorrectly translates Chinese Braille into the English word "Nine" and makes errors based on phonetic similarity. In Case 2, it introduces a spurious "\(\Rightarrow\)" at the junction of a formula and text, while also omitting characters. In contrast, BrailleLLM, with its inherent Braille knowledge, effectively mitigates such ambiguity during the translation process.

\begin{table}[!ht]
\centering
\small
\renewcommand{\arraystretch}{1.3} 
\begin{tabularx}{\columnwidth}{@{} l X @{}} 
\toprule
\multicolumn{2}{@{}l}{\textbf{Case 1: Plain text Braille translation}} \\
\midrule
\textbf{Prompt} & Please translate the following Braille into plain text: \texttt{M831 \&1+1 \#AI:GI D F9'GO1" \textbackslash\textbackslash 1 N\%\textbackslash\textbackslash 2 :1 <A I2Y :AM* D ,LOUIS ,BRAILLE SO' F9M*"2} \\
\addlinespace 
\textbf{Reference} & \begin{CJK}{UTF8}{gbsn}盲文源于19世纪的法国，由年幼时因意外失明的Louis Braille所发明。\end{CJK} \vspace{5pt} \newline  
 Braille originated in 19th-century France and was invented by Louis Braille, who lost his sight in a childhood accident. \\
\addlinespace
\textbf{Baseline} & \begin{CJK}{UTF8}{gbsn}盲文源于19世纪的法国，由{\color{red}{Nine}}时因意外{\color{red}{声明}}的LOUIS BRAILLE所发明。\end{CJK} \\
\addlinespace
\textbf{BrailleLLM} & \begin{CJK}{UTF8}{gbsn}盲文源于19世纪的法国，由年幼时因意外失明的LOUIS BRAILLE所发明。\end{CJK} \\
\midrule
\multicolumn{2}{@{}l}{\textbf{Case 2: Mixed text Braille translation}} \\
\midrule
\textbf{Prompt} & Please translate the following mixed Braille text into plain text: \texttt{GUH>'M*@ H>'G\_ZU' HO2D51 \textasciicircum:01S] H>'ZU'\textasciicircum G[L+ D ZWD9/1 W\#E16 \textbackslash" C':1 ;P*1 7;P*2 7\#A2 4} \\ 
\addlinespace
\textbf{Reference} & \begin{CJK}{UTF8}{gbsn}求解变量 y 的值，并注意定义域：\end{CJK}\texttt{\$\string\frac\{3\}\{y-2\} = \string\frac\{5\}\{y+4\}\$} \vspace{5pt} \newline Solve for the variable y, and pay attention to the domain: \texttt{\$\string\frac\{3\}\{y-2\} = \string\frac\{5\}\{y+4\}\$} \\
\addlinespace
\textbf{Baseline} & \begin{CJK}{UTF8}{gbsn}求解变量 y 的值，\textbf{\color{red}{凭注意定义}}：\end{CJK}\textbf{\color{red}{\(\Rightarrow\)}}\texttt{\$\string\frac\{3\}\{y-2\}=\string\frac\{5\}\{y+4\}\$} \\
\addlinespace
\textbf{BrailleLLM} & \begin{CJK}{UTF8}{gbsn}求解变量 y 的值，\textbf{\color{red}凭} 注意定义域：\end{CJK}\texttt{\$\string\frac\{3\}\{y-2\}=\string\frac\{5\}\{y+4\}\$} \\
\bottomrule
\end{tabularx}
\caption{Case study of Braille translation. The baseline model is Qwen2.5-7B with conventional fine-tuning. Translation errors are highlighted in red.}
\label{tab:qualitative_analysis}
\end{table}

%% file: acl_latex.bbl
\begin{thebibliography}{45}
\providecommand{\natexlab}[1]{#1}

\bibitem[{Achiam et~al.(2023)Achiam, Adler, Agarwal, Ahmad, Akkaya, Aleman, Almeida, Altenschmidt, Altman, Anadkat et~al.}]{achiam2023gpt}
Josh Achiam, Steven Adler, Sandhini Agarwal, Lama Ahmad, Ilge Akkaya, Florencia~Leoni Aleman, Diogo Almeida, Janko Altenschmidt, Sam Altman, Shyamal Anadkat, et~al. 2023.
\newblock Gpt-4 technical report.
\newblock \emph{arXiv preprint arXiv:2303.08774}.

\bibitem[{Ando and Zhang(2005)}]{Ando2005}
Rie~Kubota Ando and Tong Zhang. 2005.
\newblock A framework for learning predictive structures from multiple tasks and unlabeled data.
\newblock \emph{Journal of Machine Learning Research}, 6:1817--1853.

\bibitem[{Andrew and Gao(2007)}]{andrew2007scalable}
Galen Andrew and Jianfeng Gao. 2007.
\newblock Scalable training of {L1}-regularized log-linear models.
\newblock In \emph{Proceedings of the 24th International Conference on Machine Learning}, pages 33--40.

\bibitem[{Awang et~al.(2024)Awang, Rani, Hock, Ramly, and Kamil}]{awang2024innovative}
Azizah Awang, Intan Farahana~Abdul Rani, Kway~Eng Hock, Nur~Adillah Ramly, and Rozita Kamil. 2024.
\newblock Innovative approaches in teaching early braille reading skills: A theory and practice study.
\newblock \emph{Journal of Contemporary Social Science and Education Studies (JOCSSES) E-ISSN-2785-8774}, 4(2):177--200.

\bibitem[{Bawden et~al.(2019)Bawden, Bogoychev, Germann, Grundkiewicz, Kirefu, Barone, and Birch}]{bawden2019university}
Rachel Bawden, Nikolay Bogoychev, Ulrich Germann, Roman Grundkiewicz, Faheem Kirefu, Antonio Valerio~Miceli Barone, and Alexandra Birch. 2019.
\newblock The university of edinburgh's submissions to the wmt19 news translation task.
\newblock \emph{arXiv preprint arXiv:1907.05854}.

\bibitem[{Brown et~al.(2020)Brown, Mann, Ryder, Subbiah, Kaplan, Dhariwal, Neelakantan, Shyam, Sastry, Askell et~al.}]{brown2020language}
Tom Brown, Benjamin Mann, Nick Ryder, Melanie Subbiah, Jared~D Kaplan, Prafulla Dhariwal, Arvind Neelakantan, Pranav Shyam, Girish Sastry, Amanda Askell, et~al. 2020.
\newblock Language models are few-shot learners.
\newblock \emph{Advances in neural information processing systems}, 33:1877--1901.

\bibitem[{Cai et~al.(2019)Cai, Wang, Tang, Cui, Liu, and Qian}]{Cai2019Automatic}
Jia Cai, Xiangdong Wang, Lizhen Tang, Xiaojuan Cui, Hong Liu, and Yueliang Qian. 2019.
\newblock {Automatic Chinese-Braille Conversion Based on Chinese-Braille Contrasted Corpus and Deep Learning (in Chinese)}.
\newblock \emph{Chinese Journal of Information}, 33(4):60--67.

\bibitem[{Chowdhery et~al.(2023)Chowdhery, Narang, Devlin, Bosma, Mishra, Roberts, Barham, Chung, Sutton, Gehrmann et~al.}]{chowdhery2023palm}
Aakanksha Chowdhery, Sharan Narang, Jacob Devlin, Maarten Bosma, Gaurav Mishra, Adam Roberts, Paul Barham, Hyung~Won Chung, Charles Sutton, Sebastian Gehrmann, et~al. 2023.
\newblock Palm: Scaling language modeling with pathways.
\newblock \emph{Journal of Machine Learning Research}, 24(240):1--113.

\bibitem[{Devlin(2018)}]{devlin2018bert}
Jacob Devlin. 2018.
\newblock Bert: Pre-training of deep bidirectional transformers for language understanding.
\newblock \emph{arXiv preprint arXiv:1810.04805}.

\bibitem[{Di~Gangi and Federico(2017)}]{di2017monolingual}
Mattia~A Di~Gangi and Marcello Federico. 2017.
\newblock Monolingual embeddings for low resourced neural machine translation.
\newblock In \emph{Proceedings of the 14th International Conference on Spoken Language Translation}, pages 97--104.

\bibitem[{Dong(2023)}]{dong2023transfer}
Jun Dong. 2023.
\newblock Transfer learning-based neural machine translation for low-resource languages.
\newblock \emph{ACM Transactions on Asian and Low-Resource Language Information Processing}.

\bibitem[{Egli(2009)}]{egli2009liblouis}
Christian Egli. 2009.
\newblock Liblouis--a universal solution for braille transcription services.
\newblock In \emph{Proceedings of Daisy 2009 Conference}.

\bibitem[{Gao et~al.(2024)Gao, Hou, and Wang}]{gao2024novel}
Yuan Gao, Feng Hou, and Ruili Wang. 2024.
\newblock A novel two-step fine-tuning framework for transfer learning in low-resource neural machine translation.
\newblock In \emph{Findings of the Association for Computational Linguistics: NAACL 2024}, pages 3214--3224.

\bibitem[{Hu et~al.(2021)Hu, Hayashi, Cho, and Neubig}]{hu2021deep}
Junjie Hu, Hiroaki Hayashi, Kyunghyun Cho, and Graham Neubig. 2021.
\newblock Deep: denoising entity pre-training for neural machine translation.
\newblock \emph{arXiv preprint arXiv:2111.07393}.

\bibitem[{Huang et~al.(2023)Huang, Su, Liu, Zhang, Cai, Yuan, and Xu}]{huang2023translating}
Tianyuan Huang, Wei Su, Lei Liu, Jing Zhang, Chuan Cai, Yongna Yuan, and Cunlu Xu. 2023.
\newblock Translating braille into chinese based on improved cbhg model.
\newblock \emph{Displays}, 78:102445.

\bibitem[{Jariwala and Patel(2017)}]{jariwala2017conversion}
Nikisha~B Jariwala and Bankim Patel. 2017.
\newblock Conversion of 2d mathematical equation to linear form for transliterating into braille: an aid for visually impaired people.
\newblock \emph{Int JInnov Res Sci Eng Technol}, 6(4).

\bibitem[{Jiang et~al.(2002)Jiang, Zhu, Gielen, Dr{\'a}bek, Xia, Tan, and Bao}]{jiang2002braille}
Minghu Jiang, Xiaoyan Zhu, Georges Gielen, Elliott Dr{\'a}bek, Ying Xia, Gang Tan, and Ta~Bao. 2002.
\newblock Braille to print translations for chinese.
\newblock \emph{Information and software Technology}, 44(2):91--100.

\bibitem[{Jiang et~al.(2021)Jiang, Su, Xie, Zhouhong, Zhang, and Cai}]{Jiang2021End}
Qi~Jiang, Wei Su, Ying Xie, Anping Zhouhong, Jiuwen Zhang, and Chuan Cai. 2021.
\newblock \href {https://doi.org/10.11896/jsjkx.210100025} {{End-to-End Chinese-Braille automatic conversion based on Transformer (in Chinese)}}.
\newblock \emph{Computer Science}, 48(11A):136.

\bibitem[{Johan~Rempel and CATIS(2022)}]{johan2022importance}
CVRT Johan~Rempel and CPACC CATIS. 2022.
\newblock The importance of braille during a pandemic and beyond.
\newblock \emph{Assistive Technology Outcomes \& Benefits}, 16(2):127--134.

\bibitem[{Kana and Hagos(2024)}]{kana2024factors}
Fituma~Yadasa Kana and Asmerom~Tekle Hagos. 2024.
\newblock Factors hindering the use of braille for instruction and assessment of students with visual impairments: A systematic review.
\newblock \emph{British Journal of Visual Impairment}, page 02646196241239173.

\bibitem[{Li et~al.(2024)Li, Liu, Yan, Shao, Xie, Li, and Chi}]{li2024bilingual}
Fuxue Li, Beibei Liu, Hong Yan, Mingzhi Shao, Peijun Xie, Jiarui Li, and Chuncheng Chi. 2024.
\newblock A bilingual templates data augmentation method for low-resource neural machine translation.
\newblock In \emph{International Conference on Intelligent Computing}, pages 40--51. Springer.

\bibitem[{Li et~al.(2020)Li, Wang, and Yu}]{li2020metamt}
Rumeng Li, Xun Wang, and Hong Yu. 2020.
\newblock Metamt, a meta learning method leveraging multiple domain data for low resource machine translation.
\newblock In \emph{Proceedings of the AAAI Conference on Artificial Intelligence}, volume~34, pages 8245--8252.

\bibitem[{Liu et~al.(2024)Liu, Feng, Xue, Wang, Wu, Lu, Zhao, Deng, Zhang, Ruan et~al.}]{liu2024deepseek}
Aixin Liu, Bei Feng, Bing Xue, Bingxuan Wang, Bochao Wu, Chengda Lu, Chenggang Zhao, Chengqi Deng, Chenyu Zhang, Chong Ruan, et~al. 2024.
\newblock Deepseek-v3 technical report.
\newblock \emph{arXiv preprint arXiv:2412.19437}.

\bibitem[{Lu et~al.(2022)Lu, Zeng, Zhang, Wu, and Li}]{lu2022learning}
Yu~Lu, Jiali Zeng, Jiajun Zhang, Shuangzhi Wu, and Mu~Li. 2022.
\newblock Learning confidence for transformer-based neural machine translation.
\newblock \emph{arXiv preprint arXiv:2203.11413}.

\bibitem[{Lucas et~al.(2024)Lucas, Balad{\'o}n, Pardi{\~n}as, Ag{\"u}ero-Torales, G{\'o}ngora, and Chiruzzo}]{lucas2024grammar}
Agust{\'\i}n Lucas, Alexis Balad{\'o}n, Victoria Pardi{\~n}as, Marvin Ag{\"u}ero-Torales, Santiago G{\'o}ngora, and Luis Chiruzzo. 2024.
\newblock Grammar-based data augmentation for low-resource languages: The case of guarani-spanish neural machine translation.
\newblock In \emph{Proceedings of the 2024 Conference of the North American Chapter of the Association for Computational Linguistics: Human Language Technologies (Volume 1: Long Papers)}, pages 6385--6397.

\bibitem[{Minghu et~al.(2000)Minghu, Xiaoyan, Ying, Gang, BaoZong, and Xiaofang}]{minghu2000segmentation}
Jiang Minghu, Zhu Xiaoyan, Xia Ying, Tan Gang, Yuan BaoZong, and Tang Xiaofang. 2000.
\newblock Segmentation of mandarin braille word and braille translation based on multi-knowledge.
\newblock In \emph{WCC 2000-ICSP 2000. 2000 5th International Conference on Signal Processing Proceedings. 16th World Computer Congress 2000}, volume~3, pages 2070--2073. IEEE.

\bibitem[{Park et~al.(2020)Park, Tae, Kim, Yang, Khan, Park, and Choo}]{park2020unsupervised}
Cheonbok Park, Yunwon Tae, Taehee Kim, Soyoung Yang, Mohammad~Azam Khan, Eunjeong Park, and Jaegul Choo. 2020.
\newblock Unsupervised neural machine translation for low-resource domains via meta-learning.
\newblock \emph{arXiv preprint arXiv:2010.09046}.

\bibitem[{Qi et~al.(2018)Qi, Sachan, Felix, Padmanabhan, and Neubig}]{qi2018and}
Ye~Qi, Devendra~Singh Sachan, Matthieu Felix, Sarguna~Janani Padmanabhan, and Graham Neubig. 2018.
\newblock When and why are pre-trained word embeddings useful for neural machine translation?
\newblock \emph{arXiv preprint arXiv:1804.06323}.

\bibitem[{Radford(2018)}]{radford2018improving}
Alec Radford. 2018.
\newblock Improving language understanding by generative pre-training.

\bibitem[{Radford et~al.(2019)Radford, Wu, Child, Luan, Amodei, Sutskever et~al.}]{radford2019language}
Alec Radford, Jeffrey Wu, Rewon Child, David Luan, Dario Amodei, Ilya Sutskever, et~al. 2019.
\newblock Language models are unsupervised multitask learners.
\newblock \emph{OpenAI blog}, 1(8):9.

\bibitem[{Rasooli and Tetreault(2015)}]{rasooli-tetrault-2015}
Mohammad~Sadegh Rasooli and Joel~R. Tetreault. 2015.
\newblock \href {http://arxiv.org/abs/1503.06733} {Yara parser: {A} fast and accurate dependency parser}.
\newblock \emph{Computing Research Repository}, arXiv:1503.06733.
\newblock Version 2.

\bibitem[{S{\'a}nchez-Mart{\'\i}nez et~al.(2020)S{\'a}nchez-Mart{\'\i}nez, S{\'a}nchez-Cartagena, P{\'e}rez-Ortiz, Forcada, Espla-Gomis, Secker, Coleman, and Wall}]{sanchez2020english}
Felipe S{\'a}nchez-Mart{\'\i}nez, V{\'\i}ctor~M S{\'a}nchez-Cartagena, Juan~Antonio P{\'e}rez-Ortiz, Mikel~L Forcada, Miquel Espla-Gomis, Andrew Secker, Susie Coleman, and Julie Wall. 2020.
\newblock An english-swahili parallel corpus and its use for neural machine translation in the news domain.
\newblock In \emph{Proceedings of the 22nd Annual Conference of the European Association for Machine Translation}, pages 299--308.

\bibitem[{Sennrich(2015)}]{sennrich2015improving}
Rico Sennrich. 2015.
\newblock Improving neural machine translation models with monolingual data.
\newblock \emph{arXiv preprint arXiv:1511.06709}.

\bibitem[{Shu et~al.(2024)Shu, Zhao, Liu, Demeter, Du, and Zhang}]{shu2024lawllm}
Dong Shu, Haoran Zhao, Xukun Liu, David Demeter, Mengnan Du, and Yongfeng Zhang. 2024.
\newblock Lawllm: Law large language model for the us legal system.
\newblock In \emph{Proceedings of the 33rd ACM International Conference on Information and Knowledge Management}, pages 4882--4889.

\bibitem[{Singhal et~al.(2023)Singhal, Azizi, Tu, Mahdavi, Wei, Chung, Scales, Tanwani, Cole-Lewis, Pfohl et~al.}]{singhal2023large}
Karan Singhal, Shekoofeh Azizi, Tao Tu, S~Sara Mahdavi, Jason Wei, Hyung~Won Chung, Nathan Scales, Ajay Tanwani, Heather Cole-Lewis, Stephen Pfohl, et~al. 2023.
\newblock Large language models encode clinical knowledge.
\newblock \emph{Nature}, 620(7972):172--180.

\bibitem[{Touvron et~al.(2023)Touvron, Lavril, Izacard, Martinet, Lachaux, Lacroix, Rozi{\`e}re, Goyal, Hambro, Azhar et~al.}]{touvron2023llama}
Hugo Touvron, Thibaut Lavril, Gautier Izacard, Xavier Martinet, Marie-Anne Lachaux, Timoth{\'e}e Lacroix, Baptiste Rozi{\`e}re, Naman Goyal, Eric Hambro, Faisal Azhar, et~al. 2023.
\newblock Llama: Open and efficient foundation language models.
\newblock \emph{arXiv preprint arXiv:2302.13971}.

\bibitem[{Vaswani(2017)}]{vaswani2017attention}
A~Vaswani. 2017.
\newblock Attention is all you need.
\newblock \emph{Advances in Neural Information Processing Systems}.

\bibitem[{Wang et~al.(2016)Wang, Yang, Liu, and Qian}]{wang2016chinese}
Xiangdong Wang, Yang Yang, Hong Liu, and Yueliang Qian. 2016.
\newblock Chinese-braille translation based on braille corpus.
\newblock \emph{International Journal of Advanced Pervasive and Ubiquitous Computing (IJAPUC)}, 8(2):56--63.

\bibitem[{Wang et~al.(2017)Wang, Yang, Zhang, Jiang, Liu, and Qian}]{wang2017chinese}
Xiangdong Wang, Yang Yang, Jinchao Zhang, Wenbin Jiang, Hong Liu, and Yueliang Qian. 2017.
\newblock Chinese to braille translation based on braille word segmentation using statistical model.
\newblock \emph{Journal of Shanghai Jiaotong University (Science)}, 22:82--86.

\bibitem[{Wang et~al.(2019)Wang, Zhong, Cai, Liu, and Qian}]{wang2019cbconv}
Xiangdong Wang, Jinghua Zhong, Jia Cai, Hong Liu, and Yueliang Qian. 2019.
\newblock Cbconv: service for automatic conversion of chinese characters into braille with high accuracy.
\newblock In \emph{Proceedings of the 21st International ACM SIGACCESS Conference on Computers and Accessibility}, pages 566--568.

\bibitem[{Yu et~al.(2023)Yu, Su, Liu, Zhang, Cai, and Xu}]{yu2023pre}
HaiLong Yu, Wei Su, Lei Liu, Jing Zhang, Chuan Cai, and Cunlu Xu. 2023.
\newblock Pre-training model for low-resource chinese--braille translation.
\newblock \emph{Displays}, 79:102506.

\bibitem[{Zatserkovnyi et~al.(2019)Zatserkovnyi, Mayik, Zatserkovna, and Mayik}]{zatserkovnyi2019analysis}
RH~Zatserkovnyi, VZ~Mayik, RS~Zatserkovna, and L~Ya Mayik. 2019.
\newblock Analysis of braille translation software.
\newblock \emph{Printing and Publishing}, 2:36--44.

\bibitem[{Zhang et~al.(2022)Zhang, Chen, Chen, Chen, Zhong, and Zeng}]{zhang2022design}
Ju-Xiao Zhang, Hai-Feng Chen, Bing Chen, Bei-Qin Chen, Jing-Hua Zhong, and Xiao-Qin Zeng. 2022.
\newblock Design and implementation of chinese common braille translation system integrating braille word segmentation and concatenation rules.
\newblock \emph{Computational Intelligence and Neuroscience}, 2022(1):8934241.

\bibitem[{Zhang et~al.(2019)Zhang, Kishore, Wu, Weinberger, and Artzi}]{zhang2019bertscore}
Tianyi Zhang, Varsha Kishore, Felix Wu, Kilian~Q Weinberger, and Yoav Artzi. 2019.
\newblock Bertscore: Evaluating text generation with bert.
\newblock \emph{arXiv preprint arXiv:1904.09675}.

\bibitem[{Zhou et~al.(2022)Zhou, Meng, Zhou, Zhang, Wang, and Su}]{zhou2022confidence}
Chulun Zhou, Fandong Meng, Jie Zhou, Min Zhang, Hongji Wang, and Jinsong Su. 2022.
\newblock Confidence based bidirectional global context aware training framework for neural machine translation.
\newblock \emph{arXiv preprint arXiv:2202.13663}.

\end{thebibliography}
